\documentclass[letterpaper, 10 pt, conference]{ieeeconf}  % Comment this line out if you need a4paper

\IEEEoverridecommandlockouts                              % This command is only needed if 
\usepackage{xurl}
\usepackage{graphicx}
\usepackage{booktabs}
\usepackage{multirow}
\usepackage{caption}
\usepackage{amssymb} % for checkmarks
\usepackage[hidelinks]{hyperref}

\usepackage{fancyhdr}
\usepackage{xcolor}

\newcommand{\ieeeCitation}{%
  \textcolor{gray}{\sffamily\footnotesize
  \textbf{Published in:} Proceedings of the 2026 IEEE Conference on Artificial Intelligence (CAI). 8-10 May, 2026. Granada, Spain \\
  \textbf{DOI:} \url{https://doi.org/10.1109/CAI68641.2026.11536530}}%
}

\newcommand{\ieeeNotice}{%
  \textcolor{gray}{\footnotesize
  \textcopyright~2026 IEEE. Personal use of this material is permitted.
  Permission from IEEE must be obtained for all other uses, in any current or future
  media, including reprinting/republishing this material for advertising or promotional
  purposes, creating new collective works, for resale or redistribution to servers or
  lists, or reuse of any copyrighted component of this work in other works.}%
}

\fancypagestyle{ieeeCopyright}{%
  \fancyhf{} % Clear default headers and footers

  \fancyhead[C]{\raisebox{5mm}[0pt][0pt]{\parbox{\textwidth}{\centering\ieeeCitation}}}
  \fancyfoot[C]{%
    \parbox{\textwidth}{%
      \centering
      \renewcommand{\baselinestretch}{0.75}
      \selectfont
      \ieeeNotice
    }%
  }
}

\title{\LARGE \bf
OGD4All: A Framework for Accessible Interaction with Geospatial Open Government Data Based on Large Language Models
}

\author{Michael Siebenmann$^{1,2}$, Javier Argota Sánchez-Vaquerizo$^{1}$, \\
Stefan Arisona$^{2}$, Krystian Samp$^{2}$, Luis Gisler$^{2}$, Dirk Helbing$^{1,3}$
\thanks{$^{1}$Professorship of Computational Social Science, ETH Zurich,
        8092 Zürich, Switzerland}%
\thanks{$^{2}$Esri R\&D Center Zurich, 8005 Zürich, Switzerland}%
\thanks{$^{3}$Complexity Science Hub, Vienna, 1030, Austria}%
}

\begin{document}

\maketitle
\thispagestyle{ieeeCopyright}
\pagestyle{empty}

%%%%%%%%%%%%%%%%%%%%%%%%%%%%%%%%%%%%%%%%%%%%%%%%%%%%%%%%%%%%%%%%%%%%%%%%%%%%%%%%

\begin{abstract}
We present OGD4All, a transparent, auditable, and reproducible framework based on Large Language Models (LLMs) to enhance citizens' interaction with geospatial Open Government Data (OGD).
The system combines semantic data retrieval, agentic reasoning for iterative code generation, and secure sandboxed execution that produces verifiable multimodal outputs.
Evaluated on a 199-question benchmark covering both factual and unanswerable questions, across 430 City-of-Zurich datasets and 11 LLMs, OGD4All reaches 98\% analytical correctness and 94\% recall while reliably rejecting questions unsupported by available data, which minimizes hallucination risks.
Statistical robustness tests, as well as expert feedback, show reliability and social relevance.
The proposed approach shows how LLMs can provide explainable, multimodal access to public data, advancing trustworthy AI for open governance.
\end{abstract}

%%%%%%%%%%%%%%%%%%%%%%%%%%%%%%%%%%%%%%%%%%%%%%%%%%%%%%%%%%%%%%%%%%%%%%%%%%%%%%%%
\section{INTRODUCTION}

\addtocounter{footnote}{3} 

Allowing citizens to retrieve relevant open documents and datasets, which directly answer their specific questions, is a fundamental step towards e-governance \cite{Helbing2023DemocracySociety}. While existing Open Government Data (OGD) repositories, e.g., data.gov, data.europe.eu, and opendata.swiss, provide massive data collections, their access has remained limited by non-semantic search engines, inadequate metadata and documentation, as well as the need for specialized expertise. Often, citizens are forced to manually inspect dozens of datasets in order to verify their relevance.

%%%%%%%%%%

When citizens manage to find and retrieve their desired dataset or document, the next challenge is to analyze it.
While non-technical, text-based documents may still be interpretable by citizens, vast datasets with thousands of attributes and entries are typically not.
This issue is particularly pronounced with geospatial datasets, which often come in technical formats that demand specialized knowledge and GIS software.

In recent years, LLMs equipped with appropriate tools have demonstrated increasing proficiency in navigating large data contexts \cite{schick_toolformer_2023, lewis_retrieval-augmented_2020, yao_react_2023} and generating task-appropriate analysis code \cite{gao_pal_2023, yin_natural_2022}.
However, when directly prompted to analyze data or provide insights related to public life, current frontier LLMs have been repeatedly observed to produce hallucinated, incorrect information \cite{huang_survey_2025}. 
Even with Retrieval-Augmented Generation (RAG), the exact source underlying the output produced can be hard to find or verify.
Further, platforms like Reddit and Facebook are among the most commonly cited sources of LLMs, according to a 2025 study by Semrush.\footnote{\url{https://www.semrush.com/blog/ai-mode-comparison-study/}, last accessed: 21 August 2025} This, however, is problematic as these platforms primarily host user-generated content that is rarely subject to editorial oversight or fact-checking.

Ultimately, the challenges of citizens to effectively harness OGD has broader societal implications.
While many countries have passed laws ensuring freedom of information, in practice, these rights have little impact if the data is only accessible to a small fraction of citizens. In addition, there are further enabling conditions, such as freedom of the press and political agency \cite{peixoto_uncertain_2013}.
Widely accessible tools for extracting insights from open data can foster evidence-based political deliberation, support journalism and accountability, and enable local communities to address public issues collaboratively.

Hence, we have developed OGD4All (\textit{``Open Government Data For All''}) as a transparent, end-to-end, auditable, and reproducible LLM framework enhancing the interaction with geospatial OGD for citizens, which integrates:
\begin{enumerate}
    \item Semantic retrieval from large metadata sets with robust rejection of unanswerable queries to prevent hallucination
    \item Iterative code-generation agents executing within a secure sandbox for verifiable geospatial analysis
    \item A manually curated and parameterized 199-question benchmark to assess correctness, robustness, retrieval, recall, and costs
\end{enumerate}
Together, these components position OGD4All as a reproducible tool that supports responsible and transparent AI applications in open governance.

%%%%%%%%%%%%%%%%%

\section{RELATED WORK}

Early work on integrating LLMs with data analysis tasks focused on tabular reasoning and tool-calling.
This approach has been explored in design-science research \cite{schelhorn_designing_2024}, which combines tabular data analysis and tool calling for an OGD assistant.
Recent advances extend these capabilities to geospatial contexts, as seen in GeoGPT \cite{zhang_geogpt_2024}, UrbanLLM \cite{jiang2024urbanllm}, GeoTool-GPT \cite{wei_geotool-gpt_2025}, and related systems, combining LLMs with GIS tools.
It has been repeatedly shown that generating code is a more promising approach than calling tools, notably due to better composability, generality, and representation in LLM training corpora \cite{yang_if_2024, wang_executable_2024}.
Some GIS-focused systems harnessing these advantages have been proposed, for example, with LLM-Geo \cite{li_autonomous_2023} or ChatGeoAI \cite{mansourian_chatgeoai_2024}.

Regardless of the chosen approach, many of these systems operate on curated datasets and emphasize technical performance rather than reproducibility, transparency, or public-sector applicability.
Moreover, none of the surveyed systems incorporates built-in rejection mechanisms to prevent hallucinated answers in case of no available data.
Additionally, they lack an automated and comprehensive benchmarking framework that evaluates not only analytical correctness, but also retrieval performance and practical deployment factors such as API costs, latency, and token consumption.

In parallel to the aforementioned academic works, government initiatives have aimed to improve access to OGD through conversational and visual interfaces\footnote{\url{https://stp.wien.gv.at/wienbotwidget/widget/index/}, last accessed: 3 September 2025}\textsuperscript{,}\footnote{\url{https://buenosaires.gob.ar/innovacionytransformaciondigital/boti}, last accessed: 3 September 2025}.
However, these applications typically rely on static question-answer pairs or dialogue trees, drastically limiting their ability to handle personalized or context-specific citizen questions.
More recent OGD applications like ZüriCityGPT\footnote{\url{https://chat.zuericitygpt.ch}, last accessed: 9 October 2025} demonstrate growing civic experimentation with LLM-based municipal data assistants.
Yet, they often implement a traditional RAG pipeline and thus solely utilize textual documents rather than (geospatial) datasets that require processing, and typically lack quantitative evaluations.
Overall, research on OGD highlights the need for frameworks bridging existing concerns while advancing trustworthy and explainable AI in governance \cite{Floridi2019ASociety, Shneiderman2020BridgingPractice}.

\section{METHODOLOGY}

The OGD4All framework integrates two stages: \textit{dataset retrieval} and \textit{dataset analysis}, followed by a quantitative \textit{benchmark} assessment of both. 
Combined, these components form an end-to-end architecture designed for auditable, transparent, and reproducible human–AI natural language interaction with OGD, with a focus on geospatial data. 

\begin{figure*}[!htbp]
    \centering
    \includegraphics[width=0.8\textwidth]{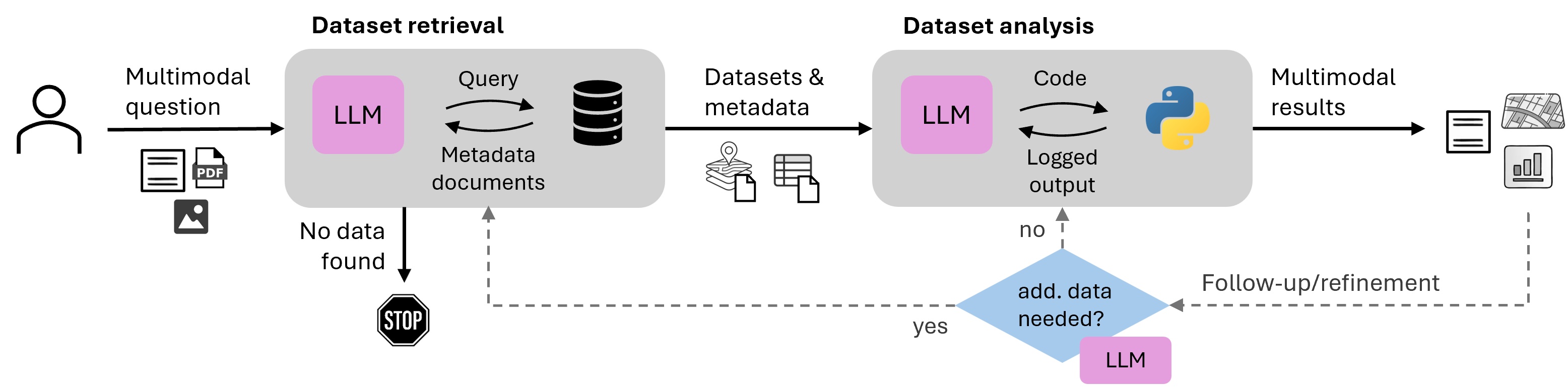}
    \caption{High-level overview of OGD4All, where an initial dataset retrieval stage is followed by a dataset analysis stage.}
    \label{fig:highLevelOverview}
\end{figure*}

Figure~\ref{fig:highLevelOverview} shows a high-level overview of the interaction with OGD4All.
An interaction starts with a citizen asking a question related to OGD, potentially including further modalities such as images or PDF documents.
In a first \textit{dataset retrieval} step, an LLM interprets the question and repeatedly queries a database containing metadata about all available datasets to identify the relevant ones for answering the question.
If no appropriate datasets are found, this is communicated to the user, and the interaction ends, reducing the risk of hallucinations.
Otherwise, each relevant dataset, accompanied by its associated metadata (field descriptions, textual information about the dataset, publication date, etc.), is passed to the \textit{dataset analysis} step.

Inside a persistent Python notebook environment, these datasets are loaded, and Python libraries required for working with geospatial data are imported.
An LLM repeatedly generates Python code and observes its outputs to work towards a solution that answers the question, until it responds with a final set of multimodal results in the form of interactive maps, plots, and text.
The citizen can ask follow-up questions or iteratively refine the system's outputs.
In this case, another LLM coordinates whether additional datasets are needed and thus appropriately routes the request either directly back to the \textit{analysis} step or to another round of \textit{retrieval}.
The process is designed to be auditable and traceable by displaying intermediate reasoning steps, generated code, and dataset sources.
This helps to ensure explainability.

\subsection{Dataset Retrieval}

In our context, the task of dataset retrieval can be defined as follows:
Given a large set of $n$ metadata documents, each containing textual information such as attributes, a summary, and the publication date of an open dataset, as well as a multimodal question (about none, one, or multiple open datasets), identify a subset of $k \leq n$ relevant datasets that can be used to answer this question.
Notably, this task definition explicitly includes questions for which no relevant datasets exist ($k=0$) or that require combining different datasets ($k>1$).
As the framework's use case is to answer questions based on open data, the case $k=0$ is used as a termination criterion.
This is an important trust mechanism usually missing in typical LLM interfaces.

A naive retrieval strategy would be to load all metadata documents into an LLM's context window and instruct it to identify the datasets relevant to a given question.
However, this approach does not scale to a large number of documents for LLMs with small context windows. Even for models with large context windows, it incurs high cost and time requirements.
Therefore, OGD4All uses a scalable approach where a metadata database is queried, and only the most relevant documents are presented to the LLM. 
A semantic retrieval strategy embeds both metadata and queries into a dense vector space and matches them using cosine similarity, thereby overcoming the vocabulary problem \cite{furnas_vocabulary_1987} common in keyword search like BM25. 
For example, a question about ``bicycles in the city'' can successfully match a document described only with the term ``urban cycling''. To this effect, OGD4All utilizes OpenAI's text-embedding-3-large model\footnote{\url{https://platform.openai.com/docs/models/text-embedding-3-large}, last accessed 29 July 2025}, which further provides multilingual embeddings.

\begin{figure}[!htb]
    \centering
    \includegraphics[width=0.5\textwidth]{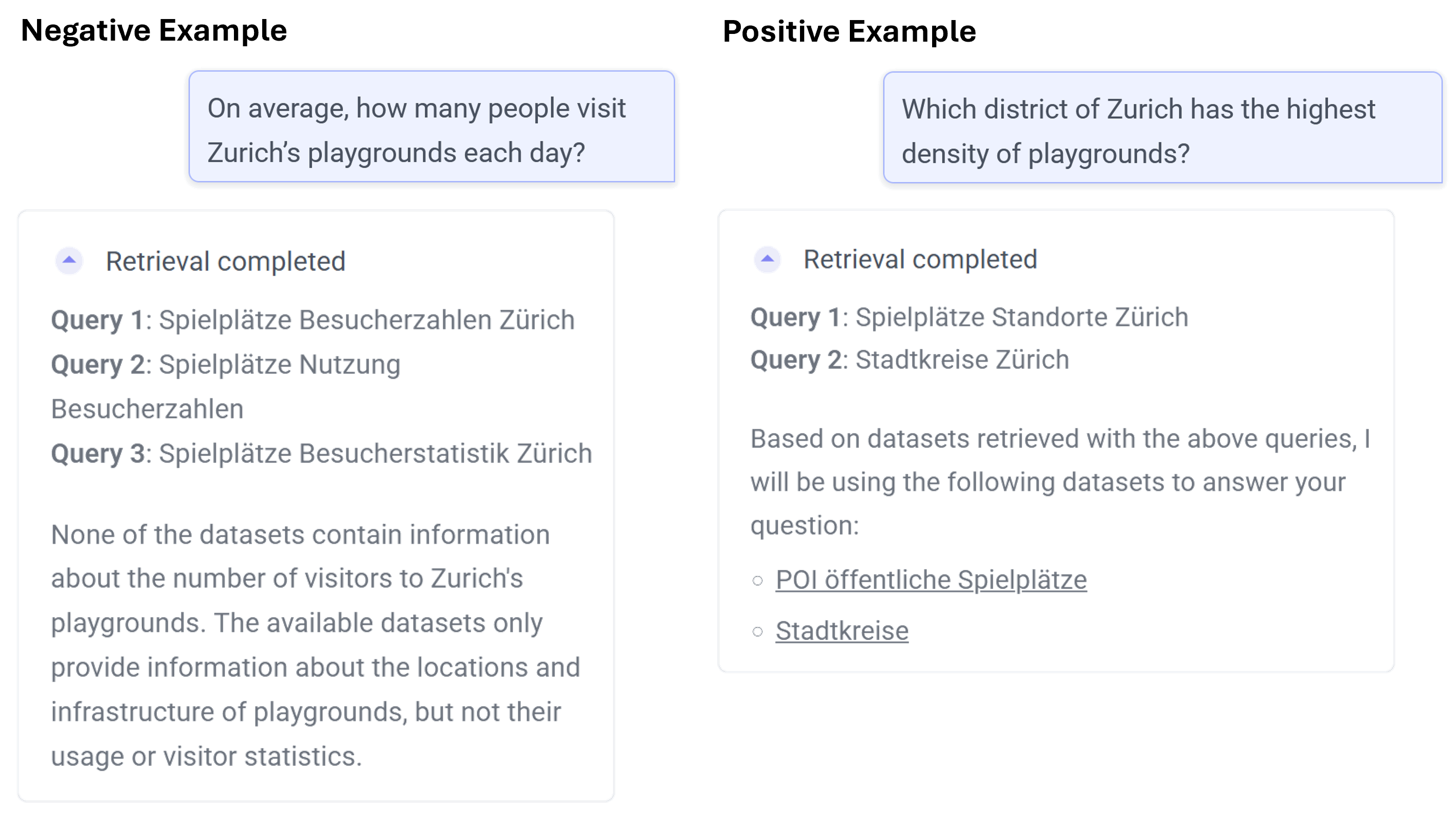}
    \caption{Screenshot of the Web-based prototype showcasing the retrieval output for two questions.}
    \label{fig:retrieval_demo}
\end{figure}

We implemented our retrieval pipeline in Python using the open-source LangGraph framework\footnote{\url{https://github.com/langchain-ai/langgraph}, last accessed: 30 July 2025} enabling agentic reasoning through structured node-based execution graphs.
Our approach relies on ``tool calling'' support, which is natively built into many recent LLMs.
``Tool calling'' lets LLMs invoke external functions, allowing them to retrieve information beyond their training data through a search tool or to perform actions such as calculations or API interactions.
In our case, the LLM is provided with two tools: one for performing a nearest neighbor search, and another for reporting results in a structured format. This allows agentic query reformulation, i.e., enabling the decomposition of the user's question into multiple focused subqueries (up to 3 by default), improving retrieval robustness.
For increased transparency, these reformulations are displayed to the user in the system's front-end.
The system prompt enabling this behavior (available in the project's repository) includes curated examples of good reformulated queries provided to the LLM (few-shot prompting). It also instructs the model to perform reformulated queries in German (and thus potentially perform translation), as all the metadata used in the prototype framework is in German. OGD4All further supports multimodal retrieval: image support for compatible models is handled by encoding as base64 and included into the LLM's conversational context, while PDF documents are converted into Markdown documents using the \texttt{pymupdf4llm} library.
The system's capability to integrate textual and visual cues is an important step towards multimodal, cognitively aligned interactions with open data.

\subsection{Dataset Analysis}
The task of dataset analysis in OGD4All is defined as follows:
Given a user's multimodal question about OGD and a set of datasets required to answer the question and corresponding metadata, generate one or multiple Python code snippets that appropriately process these datasets and produce a textual answer, as well as an interactive map, a plot, and/or a table if applicable.
This task definition focuses on the generation of code instead of an appropriate sequence of tool calls (typically in JSON format), as prior work has argued for code-based approaches to be more effective \cite{yang_if_2024, wang_executable_2024}.
Furthermore, code is more transparent and explainable than potentially ``black-box'' tool calls.
The programming language Python was chosen for this task as it consistently ranks high in LLM-oriented code generation benchmarks comparing programming languages \cite{cassano_multipl-e_2022}.
Further, it is a highly popular choice in both the GIS and Data Analysis domains, with numerous mature and comprehensive tools and libraries available.

The analysis stage is implemented through a coding agent that works towards answering the user's question step-by-step.
It does so through successive iterations of the following sub-steps: formulating a plan for the current step, generating associated code, executing it in a safe, persistent Python environment, and observing the logged output, as proposed in the CodeAct framework \cite{wang_executable_2024}.
The coding agent can recover from errors, as in such cases, the error message is provided to it, and it is instructed to generate a corrected code snippet, enabling a limited form of cognitive reflection. 
This cycle ends when the LLM agent decides to call the \texttt{final\_answer} tool, which communicates a list of results to the user, or if the maximum number of steps (20 by default) is reached.

\begin{figure}[!htb]
    \centering
    \includegraphics[width=0.5\textwidth]{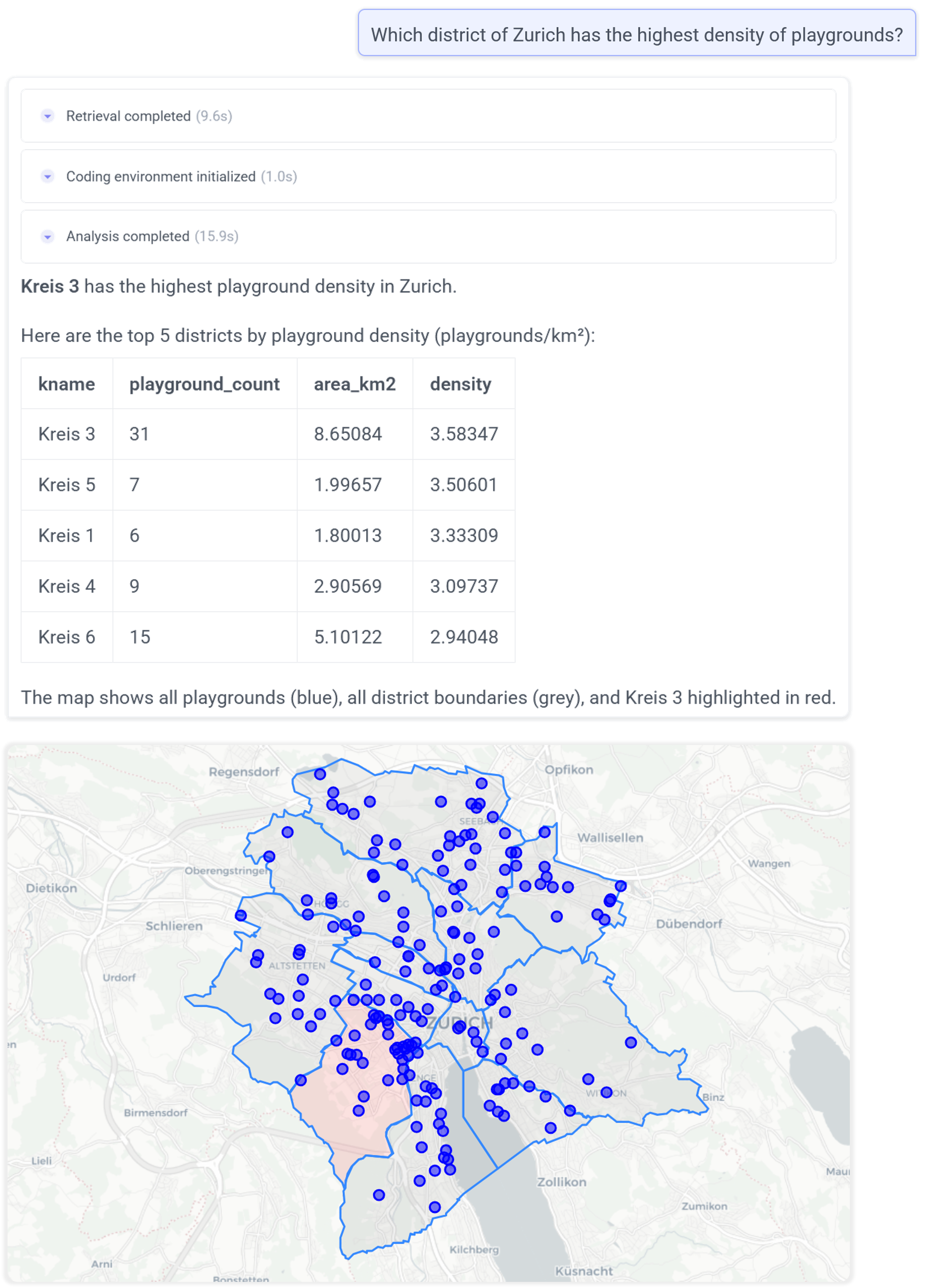}
    \caption{Screenshot of the Web-based prototype. The retrieval reformulations, links to utilized datasets, and the coding agent's step-by-step actions can be inspected by opening the accordion UI elements.}
    \label{fig:analysis_demo}
\end{figure}

As executing LLM-generated code in a local environment is inherently risky, OGD4All relies on a custom secure-by-design execution layer, namely HuggingFace's sandbox Python interpreter \texttt{smolagents}\footnote{\url{https://github.com/huggingface/smolagents}, last accessed: 04 August 2025}.
This interpreter disables additional imports and access to submodules, caps the total count of elementary operations to prevent infinite loops and resource bloating, and prevents undefined operations\footnote{\url{https://huggingface.co/docs/smolagents/tutorials/secure_code_execution\#our-local-python-executor}, last accessed: 04 August 2025}.
For maximum security (recommended for a publicly accessible deployment), OGD4All further implements support for remote code execution sandboxes using lightweight virtual machines (``\textit{microVMs}'') hosted on E2B, a cloud service provider specialized in providing secure agent environments.
However, to minimize latency and costs, the system was mainly evaluated with \texttt{smolagent}'s local Python interpreter.

\subsection{Benchmark}
To reproducibly assess the system across key metrics such as correctness, recall, and latency for different LLMs, a dedicated OGD4All benchmark that can be run with various configurations was developed. Our OGD-centric benchmark was created to cover both dataset retrieval and analysis, and it includes tasks with both geospatial and regular tabular operations over potentially multiple datasets. It features 169 manually crafted and verified question-answer pairs, from 70 parameterized templates (e.g., varying a district name or location), each accompanied by a list of relevant datasets and a ground-truth Python script for producing the answer.
To enable a robust and repeatable evaluation, the questions are precisely formulated to have an unambiguous and unique answer.
The benchmark also includes 30 negative examples, specifically questions where no answer exists based on the available data. This serves to test the system's ability to reject such inputs.
This critical ability is often overlooked in LLM benchmarks, and the commonly employed reinforcement learning based on human feedback has been shown to even worsen it, which leads to LLMs generating hallucinated answers to questions they do not have factual knowledge about \cite{liang_machine_2025}.

\begin{figure}
    \centering
    \includegraphics[width=0.5\textwidth]{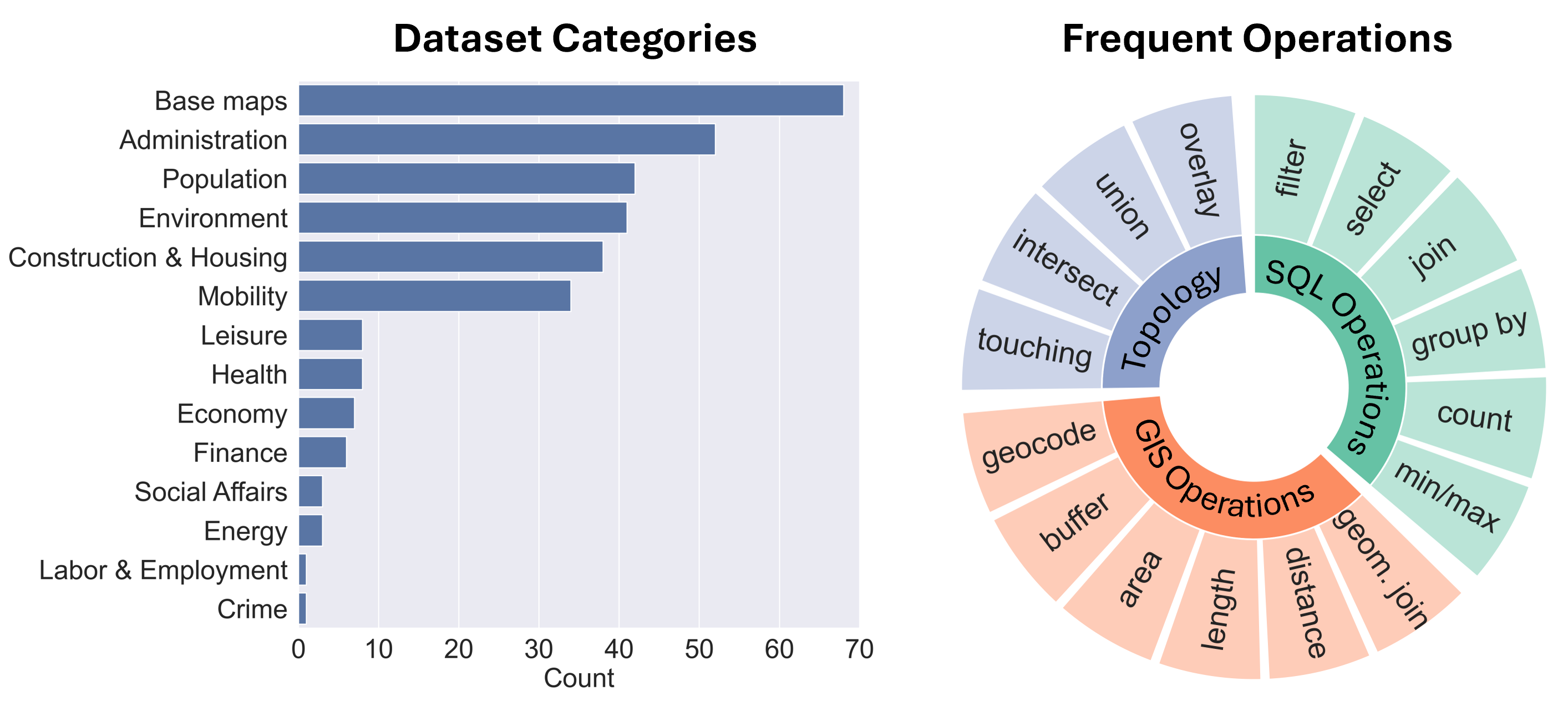}
    \caption{Benchmark dataset categories \& frequent operations}
    \label{fig:benchmark_meta}
\end{figure}

The benchmark's questions span 55 datasets from the city of Zurich's open data platform, ensuring domain realism and topical diversity.
To ensure sufficient difficulty in dataset retrieval, benchmark runs operate in a larger pool of 430 available datasets.
Among non-negative questions, 31\% require joining two or more datasets to produce the answer.
The six most frequent categories in the benchmark are base maps, administration, population, environment, construction \& housing, and mobility (see Figure~\ref{fig:benchmark_meta}), reflecting also Zurich’s OGD distribution of most frequent dataset categories.
Questions focus on geospatial open data and were, thus, created to include many typical operations in this domain.
Figure~\ref{fig:benchmark_meta} provides an overview of frequently required operations to answer benchmark questions, such as common SQL operations like filtering or aggregation, important GIS operations like geocoding or length/area computations, and topological relations such as intersections or overlays.
While the answers are limited to text in order to enable simple, automated benchmark runs, the answer types range from numerical to nominal, open, and ordinal answers.

\begin{table}[!htb]
\centering
\caption{Overview of used evaluation metrics}
\label{tab:evaluation_metrics}
\begin{tabular}{@{}p{2.3cm}p{5.9cm}@{}}
\toprule
\textbf{Metric} & \textbf{Description} \\
\midrule

Recall & Percentage of relevant datasets that were retrieved. \\
Precision & Percentage of retrieved datasets that are relevant. \\
Answerability & Accuracy of classifying whether a question can be answered with the available data or not. \\
Correctness & Whether the final answer matches the ground-truth answer. \\
Latency & Time elapsed from query submission to dataset output (retrieval) or from dataset submission to final answer (analysis). \\
Token Consumption & Total number of consumed tokens in the retrieval or analysis stage. Can be distinguished into input, output, and reasoning tokens. \\
API Cost & Total cost of retrieval or analysis stage (in USD). \\

\bottomrule
\end{tabular}
\end{table}

Table~\ref{tab:evaluation_metrics} provides an overview of the utilized metrics.
The correctness metric is computed using an \textit{LLM-as-a-judge} approach \cite{gu_survey_2025} over the 169 non-negative examples.
Although reference answers are typically concise and often comprise only a number, name, or location, this method was chosen due to the variability in answer formats and styles generated by LLMs.
Preliminary experiments demonstrated that our LLM judge based on GPT-4o is more robust in handling alternate answer representations than a traditional substring match or fuzzy match.
Recall, precision, and accuracy are reported as averages over the 199 benchmark questions.
In contrast, for latency, token consumption, and API cost, we report the median, as those output distributions were observed to be non-normal, often multimodal, or exhibiting a right skew.
Retrieval and analysis performance are assessed separately for interpretability; i.e., the ground-truth relevant datasets are supplied for the analysis benchmark runs.

We evaluate 11 closed- and open-weight LLMs 
%from/by various developers 
with regard to our benchmark.
Table~\ref{tab:llm_overview} provides an overview of the LLM selection, aimed at covering different aspects of the current LLM landscape, such as thinking models and models with a different number of parameters.
API costs are reported in USD per 1 M input/output tokens and taken from the respective official API documentations if available or from OpenRouter\footnote{\url{https://openrouter.ai}, last accessed: 1 September 2025} (as of 11 August 2025).

\begin{table}[htb]
\centering
\captionsetup{width=.8\linewidth}
\caption[Overview of evaluated LLMs]{Overview of evaluated LLMs.}
\label{tab:llm_overview}
\setlength{\tabcolsep}{4.5pt}
\begin{tabular}{l l l c l}
\toprule
\textbf{Developer} & \textbf{Name} & \textbf{\# Params} & \textbf{Thinking} & \textbf{API Cost} \\
\midrule
\multirow{6}{*}{OpenAI} 
 & GPT-4o            & undisclosed      & $\times$      & $2.5/10$      \\
 & GPT-4.1           & undisclosed      & $\times$      & $2/8$         \\
 & GPT-4.1-mini      & undisclosed      & $\times$      & $0.4/1.6$     \\
 & GPT-o1            & undisclosed      & $\checkmark$  & $15/60$       \\
 & GPT-oss 120B      & 117B/5.1B        & $\checkmark$  & $0.073/0.29$  \\
 & GPT-5             & undisclosed      & $\checkmark$  & $1.25/10$     \\
\midrule
\multirow{2}{*}{Google}
 & Gemini 2.5 Flash  & undisclosed      & $\checkmark$  & $0.3/2.5$ \\
 & Gemini 2.5 Pro    & undisclosed      & $\checkmark$  & $1.25/10$ \\
\midrule  
\multirow{2}{*}{Mistral AI}
 & Mistral Large     & 123B             & $\times$      & $2/6$         \\
 & Mistral Codestral & 22B              & $\times$      & $0.3/0.9$     \\
\midrule
\multirow{1}{*}{Meta}
 & Llama 4 Maverick  & 400B/17B         & $\checkmark$  & $0.15/0.6$      \\
\bottomrule
\end{tabular}
\end{table}

\section{RESULTS}

The following sections present the quantitative and qualitative evaluation results of OGD4All across its retrieval and analysis stages. 
Beyond accuracy and costs, our results highlight trustworthiness, transparency, and reproducibility as key dimensions of responsible human–AI collaboration in governance.
All experiments were performed using the open OGD4All benchmark, and all traces, generated code, and evaluation scripts are publicly released for verification\footnote{Documentation, source code, and supplementary information can be found on: \url{https://github.com/ethz-coss/ogd4all}}.

\subsection{Retrieval}

\begin{figure}[!htb]
    \centering
    \includegraphics[width=0.494\textwidth]{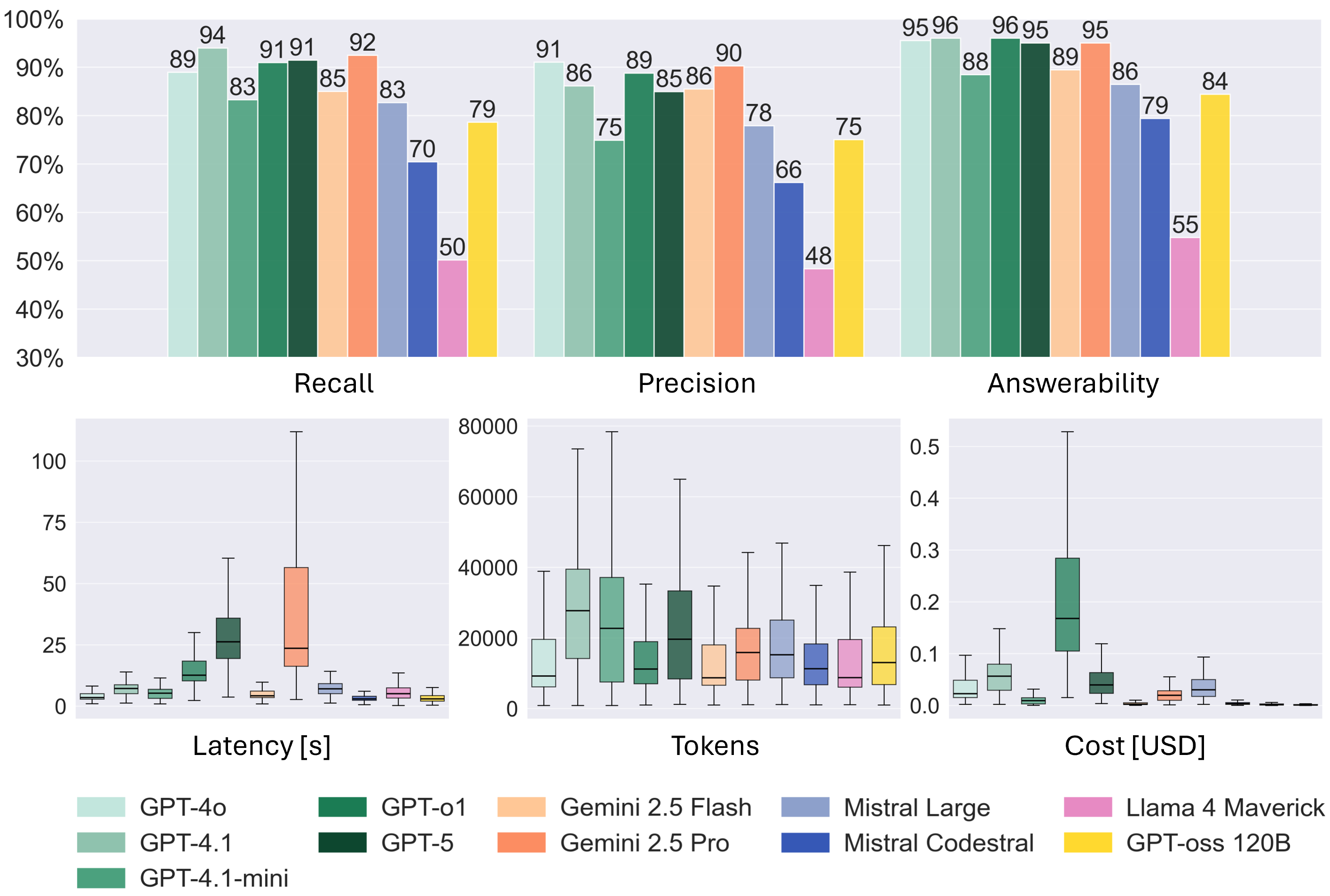}
    \caption{Retrieval metrics across different LLMs.}
    \label{fig:retrieval_llm_comparison}
\end{figure}

Figure~\ref{fig:retrieval_llm_comparison} shows benchmark results for LLM-based dataset retrieval through semantic search across 11 models.
Overall, GPT-4.1, one of OpenAI's frontier models, achieves both the highest recall and answerability accuracy scores, reaching near-perfect rejection of unanswerable questions.
These two metrics are the most critical for a trustworthy deployment, as they determine whether the system correctly grounds its responses in existing data rather than hallucinating unsupported content.
In contrast, GPT-4o scores 5\% higher in terms of precision than GPT-4.1, but with a lower recall that prevents relevant datasets from being identified, which is worse for analytical completeness. 
%%%%

As expected, one can observe that reasoning models (e.g., GPT-o1, Gemini 2.5 Pro) incur significantly higher latencies than non-reasoning variants.
Although high latency reduces usability, OGD4All's interface mitigates this by providing intermediate queries and reasoning to the user.
The differences in consumed tokens reveal consistent behavioral differences across LLMs, not only attributable to the different sizes of metadata documents (e.g., 22,000 vs.\ 200 tokens):
GPT-4.1 and its mini version employ a more exhaustive strategy and issue more refined subqueries (an average of 2.25 per question) compared to other models (1.25-1.6), suggesting a systematic self-verification behavior that enhances reliability.

\subsection{Analysis}

\begin{figure}[!htb]
    \centering
    \includegraphics[width=0.492\textwidth]{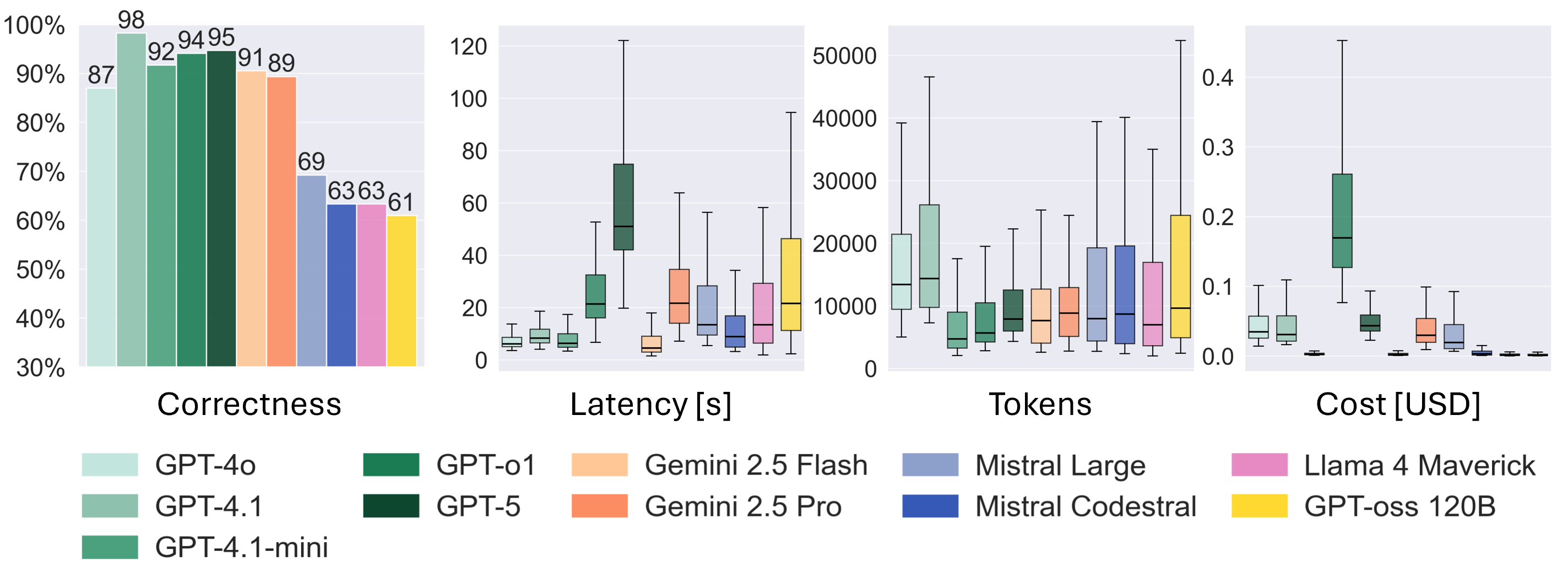}
    \caption{Analysis metrics across different LLMs.}
    \label{fig:coding_llm_comparison}
\end{figure}

Figure~\ref{fig:coding_llm_comparison} shows the analysis performance of various LLMs.
GPT-4.1 outperforms all other tested models in terms of correctness (98.2\%).
A clear performance gap is noticeable between the frontier closed-weight and open-weight models, with both Mistral models, Llama 4 Maverick, and GPT-oss 120B achieving only $60.9\%$-$69.2\%$ correctness, primarily due to lower reasoning and instruction-following abilities, and due to limited ability in generating Python code for (geospatial) data analysis.

\vspace{40pt}

GPT-4.1-mini is arguably the best choice for cost-sensitive applications, as it achieves a correctness rate of 91.7\% at 15 times lower median API costs per question compared to GPT-4.1.
The reasoning models GPT-5, GPT-o1, and Gemini 2.5 Pro exhibit significantly higher latencies than most non-reasoning models (median 51.1~s for GPT-5).
Surprisingly, GPT-oss 120B also features a high median latency of 21.6~s, attributable to this model frequently generating coding errors and thus requiring many extra iterations for error recovery.
Despite being OpenAI's most recent frontier model (as of 10 September 2025), GPT-5 shows a decline in correctness rate compared to GPT-4.1, suggesting limited benefits for short analytical tasks.

For GPT-4.1, the most common reasons for an incorrect answer are misunderstandings of the potentially insufficient data or the user intent, poor geometric approximations, and reasoning errors.
Importantly, no incorrect claims were produced through LLM-generated, hallucinated data.
Instead, generated code and reasoning traces made errors auditable and explainable, which are essential features for transparent AI.
For example, the question \textit{``What is the ratio (in percent) of disabled parking spaces to the total number of public parking spaces? Round to two decimal places.''} (ID: 6) was consistently answered with a small bias due to dataset overlap, which can be verified by inspecting the corresponding Python code.

\subsection{Expert Feedback and Deployment Considerations}

To assess a potential real-world implementation of OGD4All, we surveyed six municipal experts from the city administration of Zurich.
Each expert was informed about the system using a short, 4-minute demonstration video\footnote{\url{https://drive.google.com/file/d/1wy7I5ksHwEPdCRwhL0Wg26dcEmsBE0OJ/view}, last accessed: 05 August 2025}, followed by a brief questionnaire designed to assess their perception of its added value, potential risks, and system requirements.
While they confirmed the system's high value for citizen engagement, their main concern was the risk of generating incorrect information, which could affect public trust.
This reinforces our focus on hallucination avoidance and auditable reasoning, confirming that transparency is a prerequisite for trust in AI-driven governance systems.

Experts also pointed to a practical dilemma: the higher performance of proprietary models such as GPT-4.1 (especially in the analysis stage) conflicts with public data and governance sovereignty principles.
This underscores the key policy challenge of bridging the gap between technical performance and governance accountability through the maturation of open and sovereign LLMs.

\subsection{Limitations and Future Work}
As seen in Figures \ref{fig:retrieval_llm_comparison} \& \ref{fig:coding_llm_comparison}, current frontier models already saturate our custom benchmark with near-perfect scores in recall, answerability, and correctness.
While this limits the discriminability for future frontier models, it also clearly indicates that current models can be used for reliably solving typical, short retrieval and analysis tasks related to OGD.
Benchmark complexity could be increased by augmenting the benchmark with question-answer pairs containing domain expert knowledge collected from experts and local knowledge.
Furthermore, each benchmark question had a well-defined and unique answer.
The system's behavior for ambiguous or open questions and prompts was also not evaluated.
However, in the geospatial context, ambiguity in the formulation of a question is a common issue, as discussed in detail by Mai et al. \cite{mai_geographic_2021}.

While maps and plots are commonly produced during the analysis, the automated benchmark evaluation grades only textual outputs.
As a result, the quality of maps and plots was not quantitatively evaluated.
However, qualitative feedback from interviewed municipal experts confirmed the high value of these visualizations for citizen communication. 
This highlights the importance of developing multimodal evaluation methods in future studies.

\section{CONCLUSIONS}

The findings from OGD4All illustrate how LLMs can be operationalized for OGD interactions while promoting transparency and accountability.
Beyond emphasizing model performance, our study highlights the importance of trust, explainability, and reproducibility as design objectives for generative AI systems deployed in public governance.

%Our experiments showed that the choice of LLM directly impacts these qualities: proprietary frontier models like GPT-4.1 exhibit higher factual reliability, whereas open-weight models lag behind by reasoning limitations that affect traceability and precision. % NOTE: adapted below as I don't see how factual reliability would impact traceability? also reasoning limitations is a bit of a vague term, and the specific "root cause" error analysis was only done for GPT-4.1
Our experiments showed that the choice of LLM directly impacts these qualities: proprietary frontier models like GPT-4.1 more consistently identify relevant datasets and exhibit higher factual reliability than tested open-weight models.
This insight points to the broader research challenge of developing sovereign open models that can meet governance standards for grounded, verifiable reasoning.

Beyond benchmarking, OGD4All contributes a replicable framework for human-AI systems that is technically reliable and socially aligned.
Its logged reasoning traces, secure code execution, and open repository show that transparency-by-design can be engineered as a system property rather than an afterthought.

Future research should also extend the benchmarking towards multimodal reasoning, ambiguous query handling, and participatory validation, including domain experts.
Such directions will determine how generative AI systems evolve from experimental prototypes to trustworthy infrastructures for open and democratic governance.

%\addtolength{\textheight}{-9cm}

\section*{ACKNOWLEDGMENT}

The authors acknowledge that they were able to use the AI infrastructure of Esri to interface with various LLMs.
JASV and DH are grateful for support by the project “CoCi: Co-Evolving City Life”, which received funding from the European Research Council (ERC) under the European Union’s Horizon 2020 research and innovation programme under grant agreement No. 833168.

%%%%%%%%%%%%%%%%%%%%%%%%%%%%%%%%%%%%%%%%%%%%%%%%%%%%%%%%%%%%%%%%%%%%%%%%%%%%%%%%
%\clearpage
\vspace{40pt}
\enlargethispage{-4cm}
\bibliographystyle{IEEEtran}
\bibliography{references-2}

\end{document}